\documentclass{article}
\usepackage[preprint,nonatbib]{neurips_2020}
\usepackage[utf8]{inputenc} 
\usepackage[T1]{fontenc}    
\usepackage{hyperref}       
\usepackage{url}            
\usepackage{booktabs}       
\usepackage{amsfonts}       
\usepackage{nicefrac}       
\usepackage{microtype}      
\usepackage{graphicx}
\usepackage{mathtools}
\usepackage{wrapfig}
\usepackage{makecell}
\usepackage{appendix}
\usepackage{enumitem}

\DeclareMathOperator{\sign}{sign}

\DeclareMathOperator*{\argmax}{arg\,max}

\title{Training highly effective connectivities within neural networks with randomly initialized or constant magnitude fixed weights}

\author{
	Cristian Ivan, Răzvan Valentin Florian\\
	Romanian Institute of Science and Technology,
	Cluj-Napoca, Romania \\
	\texttt{\{ivan,florian\}@rist.ro} \\
}

\begin{document}
	
	\maketitle
	\begin{abstract}

		We present some novel, straightforward methods for training the connection graph of a randomly initialized neural network without training the weights. These methods do not use hyperparameters defining cutoff thresholds and therefore remove the need for iteratively searching optimal values of such hyperparameters. We can achieve similar or higher performances than in the case of training all weights, with a similar computational cost as for standard training techniques. Besides switching connections on and off, we introduce a novel way of training a network by flipping the signs of the weights. If we try to minimize the number of changed connections, by changing less than 10\% of the total it is already possible to reach more than 90\% of the accuracy achieved by standard training. We obtain good results even with weights of constant magnitude or even when weights are drawn from highly asymmetric distributions. These results shed light on the over-parameterization of neural networks and on how they may be reduced to their effective size.
		
	\end{abstract}

	\section{Introduction}\label{intro}
	
	The use of deep neural networks in many challenging areas of computer science 
	proved to be an indisputable success in recent years. 
	The increase in computing power enabled researchers to build ever growing
	model architectures with millions and even billions of parameters 
	for both supervised and unsupervised learning.
	Many successful applications of deep learning seem to favor large 
	neural networks with intricate architectures.
	Despite their effectiveness, many aspects of deep neural networks are not well understood. One such aspect is why over-parameterized models are able to generalize so well.
	
	A promising avenue of research towards a better understanding of deep learning 
	architectures is neural network pruning. Recent work in this direction showed 
	that large networks can be reduced to much smaller sub-networks while maintaining 
	their accuracy. It has been found that even very aggressively pruned 
	networks, with more than 95\% of the weights removed, performed almost as well 
	as the original \cite{LTH_original}.
	This used a surprisingly simple heuristic: at the end of training, 
	the weights with a magnitude below a certain threshold are set to zero, after which 
	the network is reset to its original state and retrained. Setting weights to 
	zero is functionally equivalent with their removal.
	This training and pruning procedure is repeated as long as the model maintains 
	an accuracy as large as the full network.
	Although capable of finding very sparse networks, this mechanism requires an
	iterative procedure as well as a hyperparameter --- a prior cutoff value for the threshold of the 
	weight magnitudes. This makes it  computationally expensive as well as
	prone to be sub-optimal due to the prior thresholds imposed on the weights.
	This pruning mechanism can be classified as a \textbf{pruning after training}
	approach.
	
	Other works such as \cite{wang2020picking,lee2018snip} use a 
	\textbf{pruning before training} approach, in order to save resources at training time. The end goal is to remove 
	connections such that the resulting network is sparse and the weights are efficiently trainable 
	after the pruning procedure.
	The third kind of approach is to \textbf{prune during training} \cite{Dai_2019,mostafa2019parameter}, where 
	dynamical pruning strategies are used in order to both prune and train weights
	at the same time. 
	
	The main goal behind these pruning strategies is to find sparse neural
	networks that can be trained to large degrees of accuracy by changing the weights. However, it has
	been shown by \cite{LTH_uber} that there exist pruning masks which can be 
	applied to an untrained network such that its performance is far better than 
	chance. Furthermore, \cite{ramanujan2019whats} developed an algorithm for 
	finding good pruning masks for networks with fixed, random weights. They
	found that neural networks can be trained to performances close to state-of-the-art without changing the weights but training just a pruning mask. 
	A downside of their algorithm is that, again, it requires an iterative procedure 
	to find the optimal threshold value for the criterion upon which the weight removal 
	is based.

	In this work we further reap the seemingly unreasonable effectiveness of neural networks
	with randomly initialized, fixed weights, with an approach of \textbf{pruning without training} the weights. 
	Our method is adjusting the connectivity
	graph of a randomly initialized neural network directly through back-propagation,
	without ever training the weights. 
	As a result, our approach has several advantages:
	\begin{itemize}[leftmargin=*]
		\item \textbf{No additional hyperparameters:} other than the network architecture (number of layers, nodes/filters per layer), learning rates and optimizer type, we do not use other predefined parameters.
		\item \textbf{Optimal pruning rates per layer:} as in \cite{liu2020dynamic}, but in contrast to other approaches where the pruning percentages are not adapted to each layer, our algorithm finds the optimal pruning rates for each layer through back-propagation alone.
		\item \textbf{Small additional computational cost:} since we do not use an iterative training and pruning procedure, our algorithm requires about as many computations as the standard way of training the weights of a neural network.
	\end{itemize}
	
	Besides the pruning paradigm where the connectivity graph is determined by switching connections on and off, we also introduce \textbf{a novel, alternative algorithm that trains the connectivity graph by just flipping the signs of the connections}.
	
	The code used for the experiments presented here is available at \href{https://github.com/rist-ro/training-neural-connectivities}{https://github.com/rist-ro/training-neural-connectivities}.

	\section{Pruning algorithm}\label{self_pruning_method}
	
	For each weight of the network we assign a trainable variable, $t$, which is passed to
	a masking function defined in the following way:
	\begin{equation}\label{eq:sign_relu}
	m(t)=\sign(\max(0,t))=\begin{cases}
	0, & t \leq 0 \\
	1, & t>0
	\end{cases}
	\end{equation}
	Applying this function on the tensor $\mathbf{T}$ with  elements $t \in \mathbb{R}$,  
	we obtain a binary mask $\mathbf{M}$ with elements $m$ from $\left\lbrace 0,1 \right\rbrace$.
	This mask is applied to the network connections: for a feed-forward network with $L$ hidden layers where each layer $l$ has $n_{l}$ units, the expression
	for the output of node $j$ in layer $l$ is then given by:
	\begin{equation}\label{maskedffd}
	h^{l}_{j}=\sigma\left( \sum_{i=0}^{n_{l-1}}h^{l-1}_{i}w^{l}_{ij}m(t^{l}_{ij})\right) 
	\end{equation}
	where $\sigma$ is the non-linear activation function, $h^{l-1}_{i}$ the output of a node from the previous layer,
	$w^{l}_{ij}$ are the incoming weights for the current node, and $m(t^{l}_{ij})$ is the masking function
	applied on the trainable value $t^{l}_{ij}$ associated with each weight $w^{l}_{ij}$. Note that we do not use
	biases in this approach. 
	
	For each forward pass a connection between
	two nodes can be enabled or disabled based on the output of $m(t)$ and is calculated automatically by the 
	network, without the need to use ad-hoc heuristics. 
	For the backward pass the gradient of the masking function is not
	defined and to overcome this issue we use the straight 
	through estimator \cite{STE}. This (biased) estimator was first proposed by Hinton
	\cite{hinton_STE} and treats the gradient of a hard threshold 
	function as if it was the identity function. Therefore the gradient is always 1.
	Unlike the approach of magnitude based pruning
	where a connection once removed it is unable to ''grow back'', with this method connections are 
	dynamically added or removed depending on how well the network performs.
	
	When we initialize the network, each layer's weights, $\mathbf{W}$,
	are drawn from a distribution $\mathcal{D}$ and the associated values, 
	$\mathbf{T}$, from a uniform distribution in the interval $(0, 0.1]$. 
	During training, the network weights are kept fixed and only $\mathbf{T}$ is updated via back-propagation.
	Given that $\mathbf{T}$ is initially strictly positive, the masks associated with the weights are initially 1 everywhere,
	effectively creating a network which uses all weights in the first forward pass.

	\subsection{Loss function}\label{lossfunction}
	
	We used both a straightforward pruning method, which we call \textit{free pruning}, as well as a method where we try to minimize the number of pruned weights while training the network, which we call \textit{minimal pruning}.
	
	For \textit{free pruning}, we minimize the loss
	\begin{equation}\label{free_loss}
	\mathcal{L}_\textrm{free}=\frac{1}{N}\sum_{i=0}^{N}L((x_i,y_i);\textbf{W})
	\end{equation}
	where $N$ is the number of samples and $L$ is the categorical cross-entropy loss. In this case, the network 
	reduces the amount of connections between layers as much as needed in 
	order to minimize the loss and therefore finds the optimal pruning
	factor for each layer. This eliminates any biased priors for the amount of pruning, which were present in other works, e.g.\ 
	\cite{LTH_original} and \cite{ramanujan2019whats}, where the pruning
	factor is specified explicitly for each layer.
	
	For \textit{minimal pruning}, we added a regularization term such that we can 
	also minimize the amount of weights that the network prunes:
	\begin{equation}\label{custom_loss}
	\mathcal{L}_\textrm{minimal}=\mathcal{L}_\textrm{free} -  \frac{1}{M}\sum_{i=0}^{M}m(t_i)
	\end{equation}
	where $M$ is the number of weights in 
	the network. In this case the regularization term keeps the number of non-zero components in the mask as high as possible
	since it essentially counts the average number of $1$'s in the masks.
	Therefore the network is constrained to prune as few weights as possible while concomitantly minimizing $\mathcal{L}_\textrm{free}$. 
	This allows us to investigate what would be a minimum amount of weights to be removed from a randomly initialized network such that it is
	still able to achieve a good performance.

	%

	\subsection{Experiments}\label{experiments_pruning}
	We have run experiments on MNIST \cite{mnist} using a LeNet-300 \cite{lenet300} architecture and CIFAR-10 using three variations of a VGG-like network \cite{vgg} as well as ResNet-18 \cite{resnet}. The network architectures are listed in table \ref{tab:table_archdetails} and we refer to them as: LeNet, Conv2, Conv4, Conv6 and ResNet. For the weight initialization we have used the two popular distributions Glorot Normal \cite{Glorot2010UnderstandingTD} and He Normal \cite{He2015DelvingDI} as well as the Signed He Constant distribution, as used by \cite{ramanujan2019whats}. For this latter distribution, each weight of a layer is set to a constant value, $\sqrt{2/n_{l-1}}$ (where $n_{l-1}$ is the number of nodes in the previous layer), and its sign is chosen randomly. Given that for standard training techniques the bias nodes are initialized to zero and then iteratively trained, in our setup we do not use biases at all, functionally equivalent to keeping them at zero. The optimizer used throughout all experiments is Adam \cite{adam}.
	
	\begin{figure}
		\centering
		\includegraphics[width=1\linewidth]{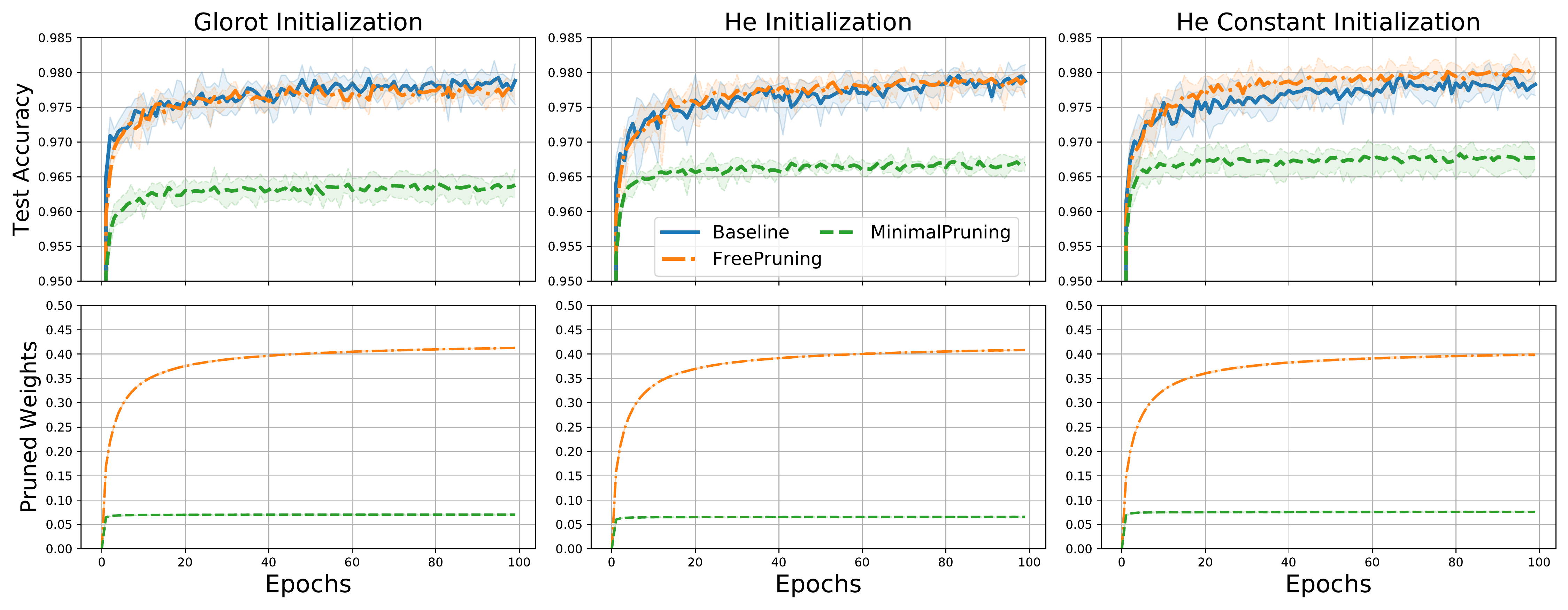}
		\caption{Pruning experiments with LeNet on MNIST with different weight initializations.}
		\label{fig:mlp_baseline_freepruning_minpruning}
	\end{figure}
	
	Our goal is to understand to what extent random weight initialization is sufficient for
	constructing highly accurate sub-networks within larger ones. In order to isolate
	the effect of pruning we train LeNet and Conv2, Conv4, and Conv6 with a minimal setup: no data augmentation, batch normalization nor any regularization techniques which may interfere with the randomness of the initialization. ResNet-18 is treated separately with a more detailed description shown in section \ref{resnet}.

	Figure \ref{fig:mlp_baseline_freepruning_minpruning} (top row panels) shows the accuracy 
	of LetNet trained on MNIST  with weights initialized from the three different distributions.
	In blue we show the baseline performance, training all weights of the network.
	In orange/green are the curves for the free/minimal pruning algorithms.
	In every experiment throughout this paper each data-point is the average of 5 
	runs and the shaded area is the minimum and maximum of the 5 runs.
	The bottom row panels (corresponding colors) indicate the fraction of pruned weights 
	as a function of the training epoch. For the baseline network, all weights are changed at each iteration,
	therefore we omitted the curve in the bottom panels.  
	One can observe that the randomly initialized network trained through free pruning reaches
	almost the same accuracy as the fully trained network. This is true for all types of
	weight initializations. The fraction of removed connections is about 45\%.

	
	Among the used distributions for the initial weights, the He Constant distribution yields the best results. With this distribution, minimal pruning achieves an accuracy of 96.8\% compared to 97.9\% for the baseline, even though the amount of 
	pruned weights is less than 8\%, while the rest of 92\% are randomly generated and remain untrained.
	\begin{figure}
		\centering
		\includegraphics[width=1\linewidth]{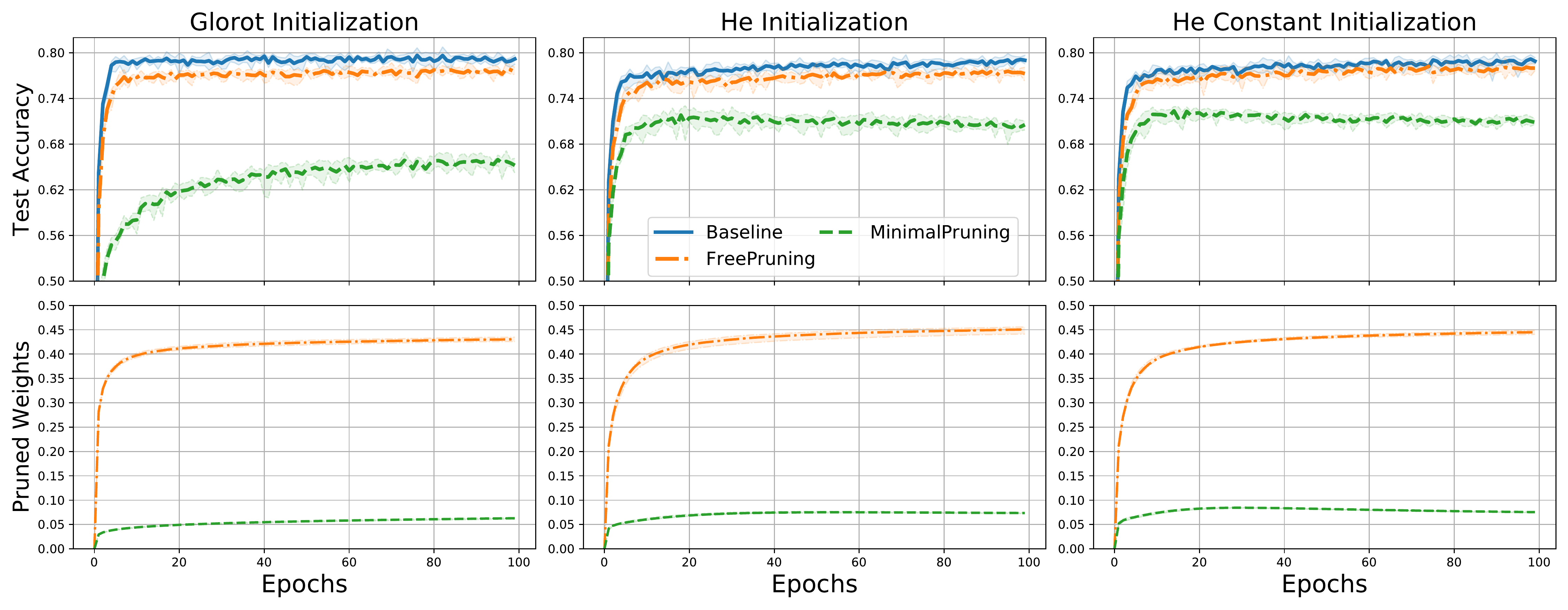}
		\caption{Pruning experiments with Conv6 on CIFAR10 with different weight initializations.}
		\label{fig:conv6_baseline_freepruning_minpruning}
	\end{figure}
	
	Figure \ref{fig:conv6_baseline_freepruning_minpruning} shows a similar behaviour 
	also when training a convolutional network on CIFAR-10.
	The randomly initialized network trained through free pruning reaches
	almost the same accuracy as the fully trained network. Minimal pruning in this case
	achieves an accuracy of about 72.3\% compared to 79.6\% for the fully trained network
	with less than 10\% of the weights removed. A similar conclusion can be drawn as in the
	case of LeNet trained on MNIST -- random weights are well enough suited for large performance.

	\subsection{Weight removal}\label{weightremoval}
	In general the forward propagation through a neural network's layer is a non-linear activation
	function of the weighted sum of the nodes from the previous layer. The formal expression for
	this is given by Eq. (\ref{maskedffd}).
	Consider the special case when we draw weights from the He Constant distribution.
	In each layer the weights are set to have the same magnitude and a randomly 
	chosen sign: $w^{l}_{ij}=\left|w^{l}\right|s^{l}_{ij}$ where $s^{l}_{ij}$ is the sign of the weight. 
	It follows that the weights from Eq. (\ref{maskedffd}) can be factored out
	of the sum. Furthermore, if we choose ReLU as the activation function then
	we have the convenient property that for any $a>0$ the function 
	$\max(0,a \mathbf{x})=a\max(0,\mathbf{x})$. 
	Equation (\ref{maskedffd}) can then be
	rewritten as:
	\begin{equation}\label{eq:hlj_factored_w}
	h^{l}_{j}=\left|w^{l}\right| \cdot \sigma\left( \sum_{i=0}^{n_{l-1}}h^{l-1}_{i} s^{l}_{ij} m(t^{l}_{ij})\right) 
	\end{equation}
	The above equation is recursively applied and as such the weights can be factored out of each layer.
	Therefore we can initialize a neural network by setting all weights to $1$, choosing a sign
	randomly and scaling the training data by the product of each layer's weight magnitude, i.e.\ $\sqrt{2/n_{l-1}}$. 
	More precisely, for a network with $L$ hidden layers where each layer has $n_{l}$ nodes, the
	training data $X$ becomes:
	\begin{equation}\label{eq:Xscaled}
	\hat{X}=X\cdot \prod_{l=1}^{L}\left|w^l\right|=X\cdot \prod_{l=1}^{L}\sqrt{\frac{2}{n_{l-1}}}
	\end{equation}
	with $n_{0}$ being the number of incoming connections from layer zero (the input data). This expression is 
	general and can be applied for fully connected layers as well as convolutional layers. Training
	the mask of a network with constant weights per layer is completely equivalent with training the mask
	of a network with unitary weights and input data scaled appropriately. 
	A result of this training technique is that
	at inference time the magnitude of the weights becomes irrelevant to the classification accuracy 
	because the output nodes are scaled by the same $W$ and we are only interested in the node with
	the highest value: $\argmax(a\mathbf{x})=\argmax(\mathbf{x})$ for any $a>0$.
	
	\begin{figure}
		\centering
		\includegraphics[width=1\linewidth]{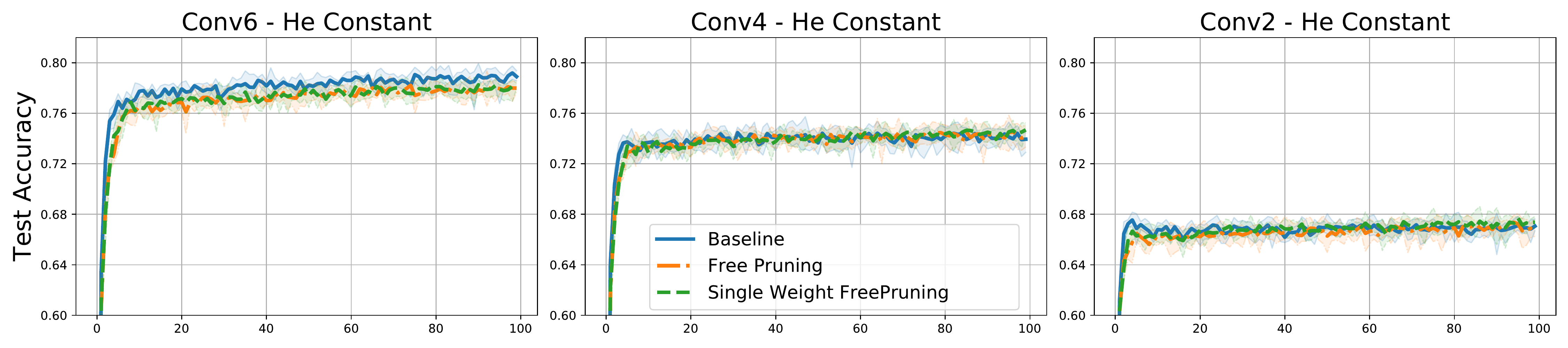}
		\caption{Conv2, Conv4 and Conv6 on CIFAR10 with He Constant initializations.}
		\label{fig:0accbaselinebinaryfreepruningc642}
	\end{figure}
	
	In order to verify that this procedure is numerically stable we have repeated the 
	same experiments described in the previous section with the new initialization scheme. 
	The results for the Conv2, Conv4 and Conv6 networks are shown in Figure \ref{fig:0accbaselinebinaryfreepruningc642}.
	We compare the accuracy of the standard procedure, where we train all weights, to
	the free pruning algorithm, with and without weight removal. The figure shows that the 
	accuracy curves of the free pruning algorithm, with and without weight removal, are almost identical in each case, and also almost identical to the ones of the baseline, for Conv2 and Conv4.

	\subsection{Weight sign imbalance}\label{weightsignimbalance}
	
	Standard weight initialization procedures use symmetric distributions: standard normal, 
	truncated normal (Glorot, He) etc. The special He Constant distribution is symmetric 
	but bimodal, having only two values, $\pm\sqrt{2/n_{l-1}}$, with the sign of the weight 
	chosen randomly from a uniform distribution.
	Drawing weights from this distribution is essentially a 
	Bernoulli trial where a weight has $p=0.5$ probability of being positive and $q=1-p=0.5$ 
	of being negative. The amount of negative/positive weights follows a binomial distribution
	and due to the large amount of parameters in standard neural networks it is extremely 
	unlikely that there is a significant imbalance between the two. 
	
	\begin{figure}
		\centering
		\includegraphics[width=.495\linewidth]{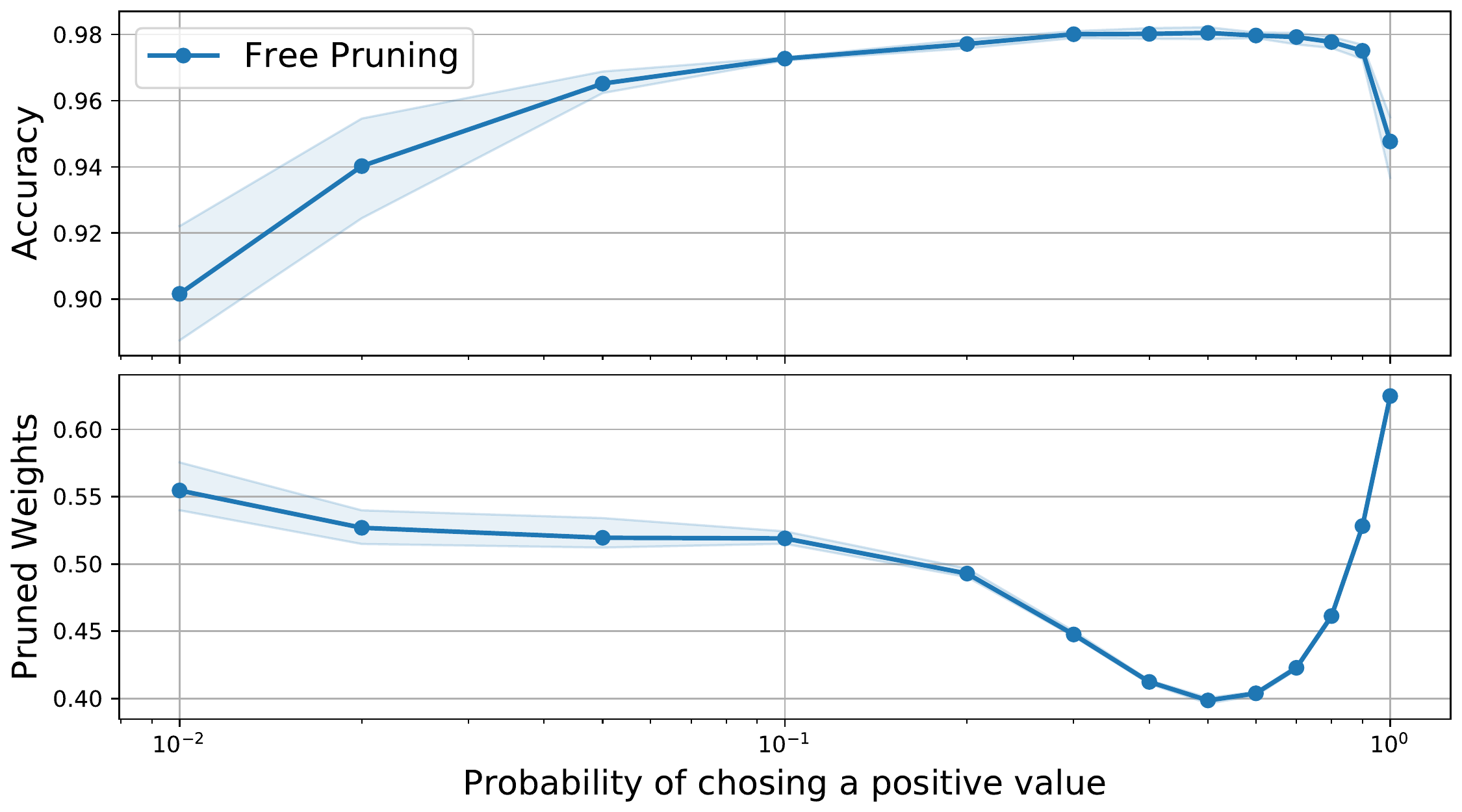}
		\includegraphics[width=.495\linewidth]{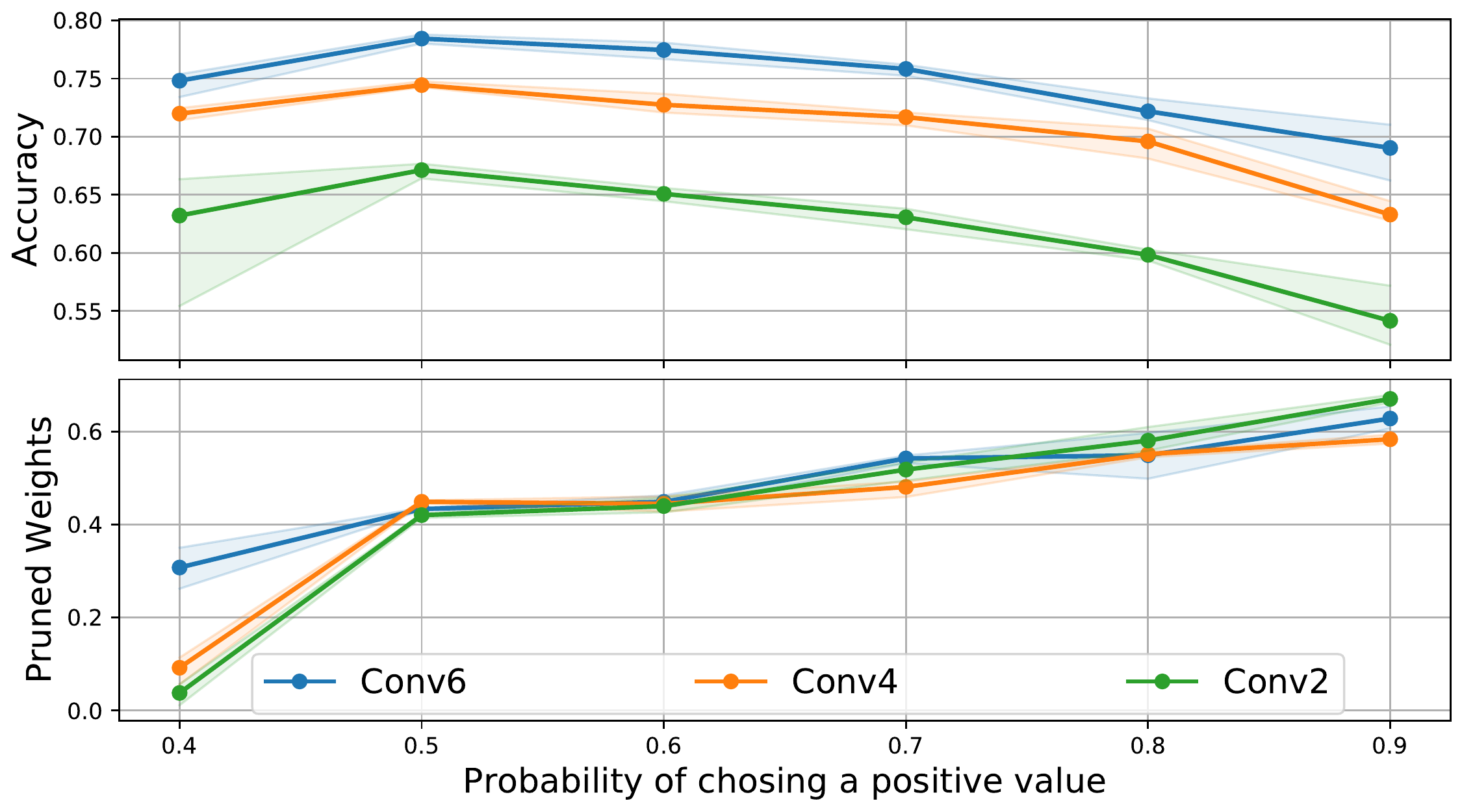}
		\caption{LeNet (left) and Conv2, Conv4, Conv6 (right) accuracies for networks with imbalanced amount of negative and positive weights. Please note the different horizontal scales, with a logarithmic one on the left and a linear one on the right.}
		\label{fig:p1_scan}
	\end{figure}
	
	We have varied the probability $p$ of obtaining a positive weight and experimented
	with different values. We obtain therefore networks with a significant imbalance
	between the number of negative and positive weights.
	Figure \ref{fig:p1_scan} (left panels) shows the dependence on $p$ of the accuracy of LeNet trained on MNIST and of the fraction of pruned weights. 
	The network maintains its performance when $0.3\leq p\leq 0.7$.
	The accuracy drops by 1\% only when reaching $p\leq0.1$ and by 10\% when $p\leq0.01$. 
	A notable result is the extreme case when $p=1$, which corresponds to a network where 
	there are only positive weights between neurons. The network is still trainable and 
	reaches 95\% accuracy with about 65\% of the weights pruned.
	In the case of Conv2, Conv4 and Conv6, right panels in Figure \ref{fig:p1_scan},
	we observe a similar behaviour as for LeNet: high accuracies are 
	reached even when initializing weights 
	from asymmetric distributions. However, convolutional networks are more sensitive
	to large asymmetries in the sign distributions. 

	\section{Sign flipping}
	
	\begin{figure}[ht]
		\centering
		\includegraphics[width=.5\linewidth]{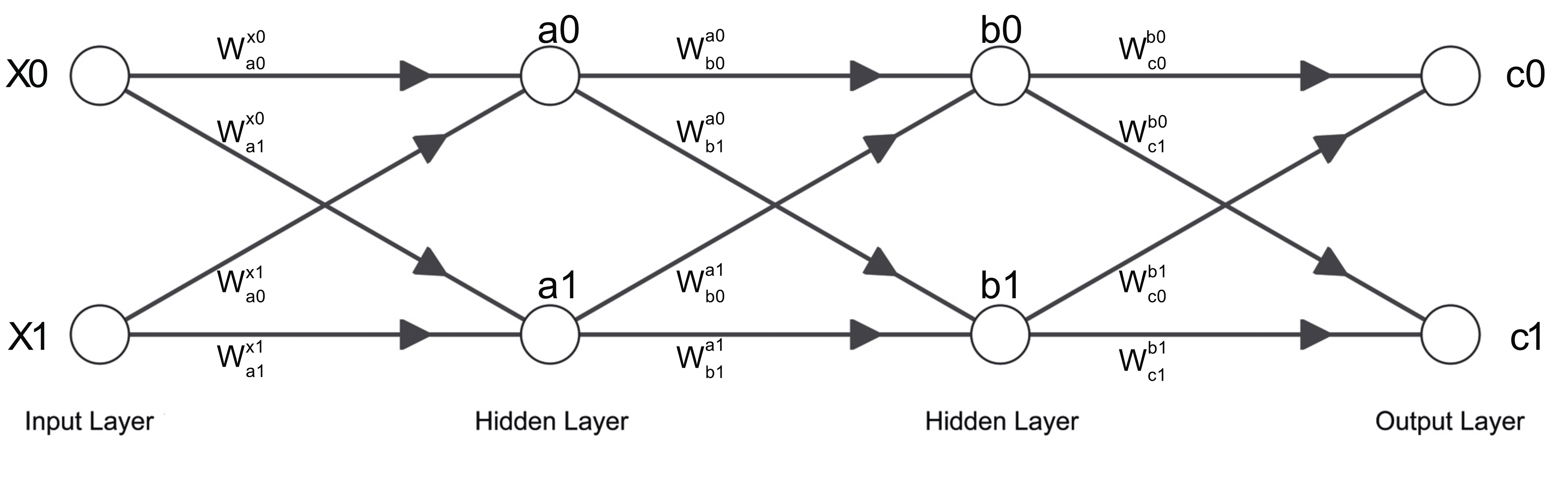}
		\caption{Simple neural network without bias nodes.}
		\label{fig:nn_1}
	\end{figure}
	
	Consider a small standard neural network with no biases as shown in Figure \ref{fig:nn_1}.
	For this simple setup one can easily work out the 
	full propagation through the network. For a given node, e.g.\ $a_0$, 
	the forward propagation is written as  $a_0=\sigma\left(x_{0}W_{a_0}^{x_0} +x_{1}W_{a_0}^{x_1}\right)$ 
	where $\sigma$ is the activation function. 
	If we choose ReLU as an activation function then we have the advantage that it can also be 
	written as $\sigma(x)=xH(x)$ where $H$ is the unit step function.
	With this in mind we can rewrite $a_0=\left(x_{0}W_{a_0}^{x_0} +x_{1}W_{a_0}^{x_1}\right)H_{a_0}$
	where we used $H_{a_0}$ as a short notation for the step function applied at $a_0$.
	It can be shown that an output node, e.g.\ $c_0$, can be expressed as:
	
	\begin{alignat}{2}\label{eq:c0_final}
	c_{0} = & x_0 \Big[ & W_{a_0}^{x_0}W_{b_0}^{a_0}W_{c_0}^{b_0}H_{a_0}H_{b_0} &+W_{a_1}^{x_0}W_{b_0}^{a_1}W_{c_0}^{b_0}H_{a_1}H_{b_0}  +\\ \nonumber
	&& W_{a_0}^{x_0}W_{b_1}^{a_0}W_{c_0}^{b_1}H_{a_0}H_{b_1} &+W_{a_1}^{x_0}W_{b_1}^{a_1}W_{c_0}^{b_1}H_{a_1}H_{b_1} \Big] +\\ \nonumber
	& 	  x_1 \Big[ & W_{a_0}^{x_1}W_{b_0}^{a_0}W_{c_0}^{b_0}H_{a_0}H_{b_0} &+W_{a_1}^{x_1}W_{b_0}^{a_1}W_{c_0}^{b_0}H_{a_1}H_{b_0}  +\\ \nonumber
	&& W_{a_0}^{x_1}W_{b_1}^{a_0}W_{c_0}^{b_1}H_{a_0}H_{b_1} &+W_{a_1}^{x_1}W_{b_1}^{a_1}W_{c_0}^{b_1}H_{a_1}H_{b_1} \Big]  
	\end{alignat}

	We can observe the following pattern: the value of the output node depends on the input values 
	$x_0$ and $x_1$, each multiplied by the sum of terms corresponding to all paths which connect these nodes.  
	Each path has an associated coefficient $H$, which is either $0$ or $1$, depending on the weighted
	sum at each intermediate node along the path. A more detailed derivation of 
	the equations in this section is given in appendix \ref{sumofpaths_derivation}.
	
	We will apply this formalism to our constrained model where weights of the neural network 
	are drawn from the Signed He Constant distribution, i.e.\ each layer has its own $\pm w_l$ value for the weights. This means that each term in Eq. (\ref{eq:c0_final}) corresponding to a path has essentially the same magnitude but with a different sign. We can replace the product of all three weights along a path with $W$ and pull out the sign of the product in a separate factor $S_{i,j,k}$ which represents the sign of the path from input node $x_i$ to output node $c_0$ through the  intermediate hidden nodes $a_j$ and $b_k$. As such $W_{a_j}^{x_i}W_{b_k}^{a_j}W_{c_0}^{b_k}=\left|W\right|S_{i,j,k}$.
	The value of $W$ is determined by the network  architecture as shown in section  \ref{weightremoval}. Rewriting all terms in the previous equation accordingly we obtain:
	
	\begin{equation}\label{eq_c01}
	\begin{split}
	c_0=&Wx_{0}\left[ S_{000}H_{a_0}H_{b_0} + S_{010}H_{a_1}H_{b_0} +S_{001}H_{a_0}H_{b_1} + S_{011}H_{a_1}H_{b_1} \right] +\\
	&Wx_{1}\left[ S_{100}H_{a_0}H_{b_0} + S_{110}H_{a_1}H_{b_0} +S_{101}H_{a_0}H_{b_1} + S_{111}H_{a_1}H_{b_1} \right]
	\end{split}
	\end{equation}
	
	The above equation can be extended for networks with arbitrary depth and width where
	the number of terms scales with the product of the number of nodes in each layer. 

	Empirically we know that, in general, many weights can be set to zero while keeping the network performance at the same level. 
	If we consider a 
	network with a fixed set of removed connections we have essentially a network with many paths of zero 
	contribution to the sum in Eq. (\ref{eq_c01}). One can observe that the contribution of sets of paths can also become zero if the signs of $S_{i,j,k}$ are carefully chosen such that the sum of terms in these sets
	is effectively zero. Therefore, we hypothesise that, analogously to how pruning can be used to train networks by setting to zero the values of some paths in Eq. (\ref{eq_c01}), networks can also be trained by changing the signs of paths, since such changes may lead to some paths cancelling the contributions of others, which is equivalent to pruning weights associated to all these sets of paths.
	
	\begin{figure}
		\centering
		\includegraphics[width=1\linewidth]{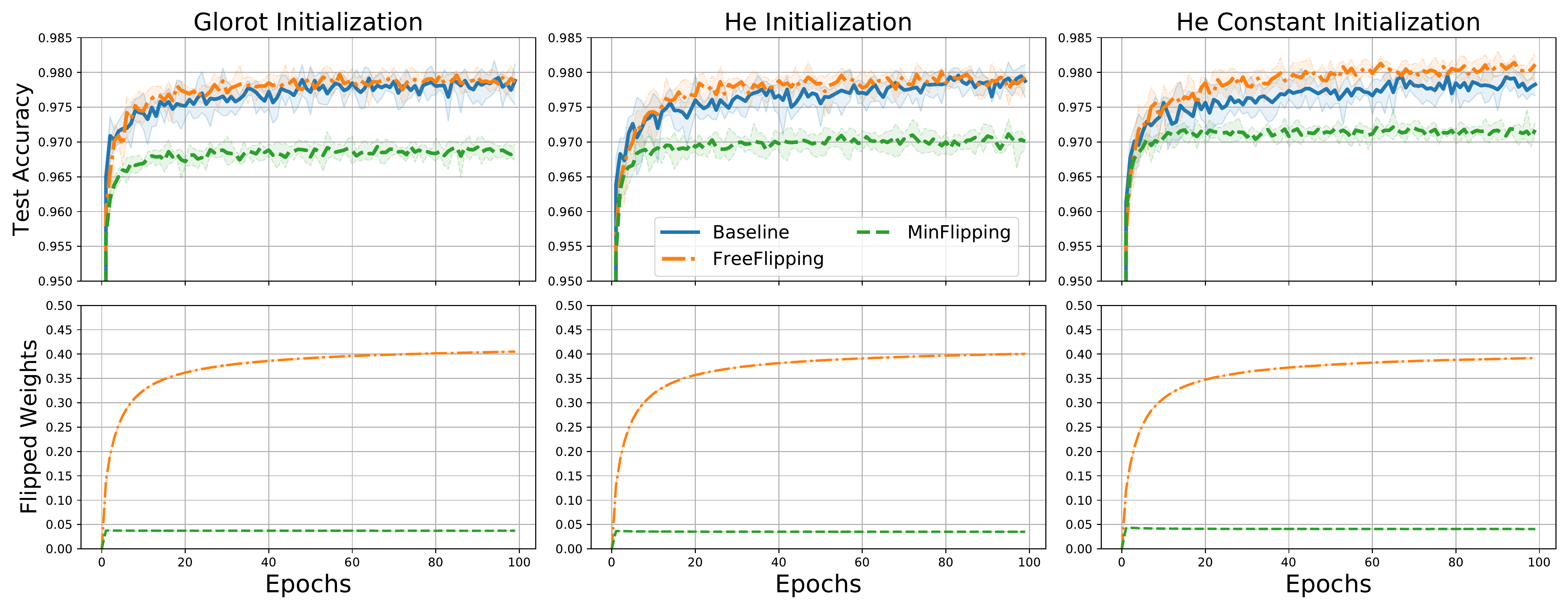}
		\caption{Sign flipping experiments with LeNet on MNIST with different weight initializations.}
		\label{fig:0accsignflipminflip}
	\end{figure}
	
	\begin{figure}
		\centering
		\includegraphics[width=1\linewidth]{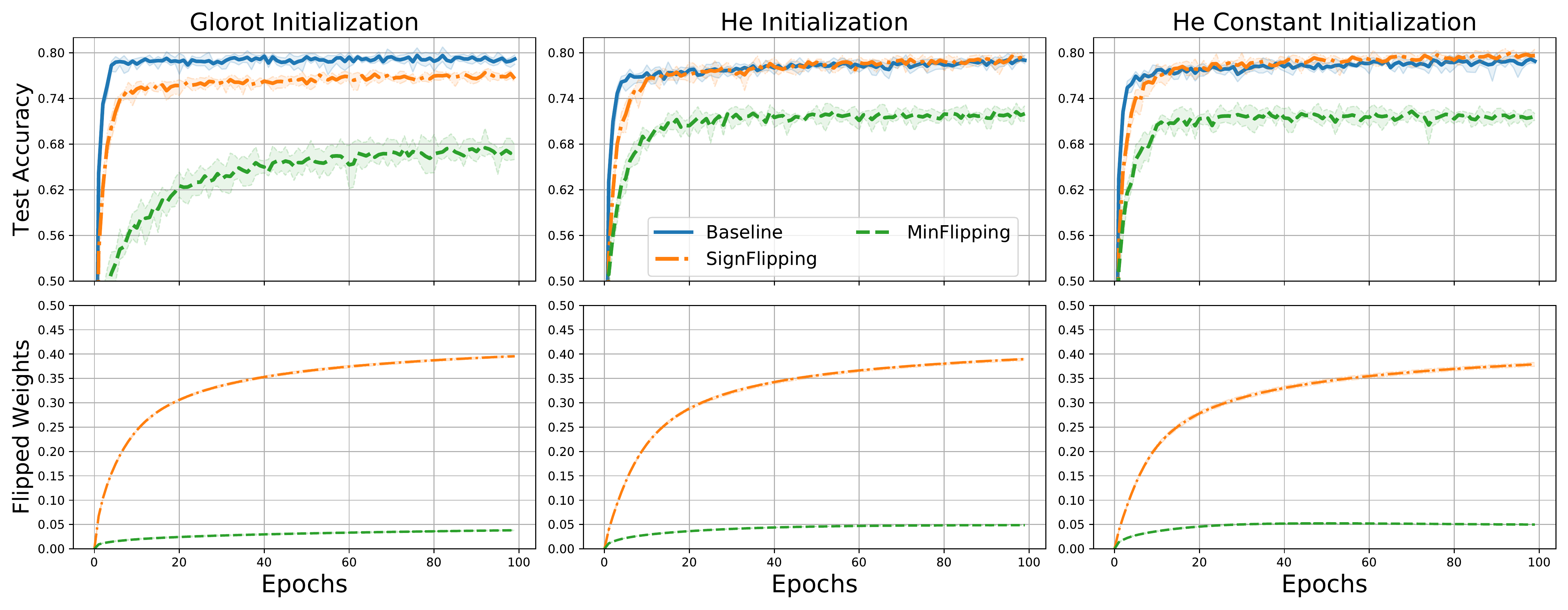}
		\caption{Sign flipping experiments with Conv6 on CIFAR10 with different weight initializations.}
		\label{fig:0accbaselinesignflipminflipc6}
	\end{figure}
	
	We have verified this hypothesis by replacing the masking function $m(t)$ 
	with a simpler function, $f(t)=\sign(t)$, such that instead of a binary mask 
	$\mathbf{M}$ with elements from $\left\lbrace 0,1 \right\rbrace$ we obtain a filter 
	$\mathbf{F}$ with elements from $\left\lbrace -1,1 \right\rbrace$ which flips the sign of a weight.
	The training procedure is the same as before: the weights are kept at their fixed, randomly initialized values and we train only a variable $t$ which is passed to the masking function. By analogy to the pruning technique, we perform either \textit{free flipping} where the network is flipping as many signs as needed in order to minimize the cross-entropy loss, or \textit{minimal flipping} where a regularization term is added in order to constrain the network to flip as few weight signs as possible.

	Figure \ref{fig:0accsignflipminflip} 
	shows the performance achieved for LeNet-300 on MNIST. We found that flipping the signs of the
	weights drawn from all three distributions works at least as good as when
	training all weights of the network in a classical manner. It is also superior to
	the pruning mechanism in both variants, free pruning as well as minimal pruning.
	For Conv6 (see Figure \ref{fig:0accbaselinesignflipminflipc6}) the accuracy is equal to or higher than the baseline for the He and He Constant initializations.

	\section{ResNet-18}\label{resnet}
	
	\begin{figure}[h]
		\centering
		\includegraphics[width=1\linewidth]{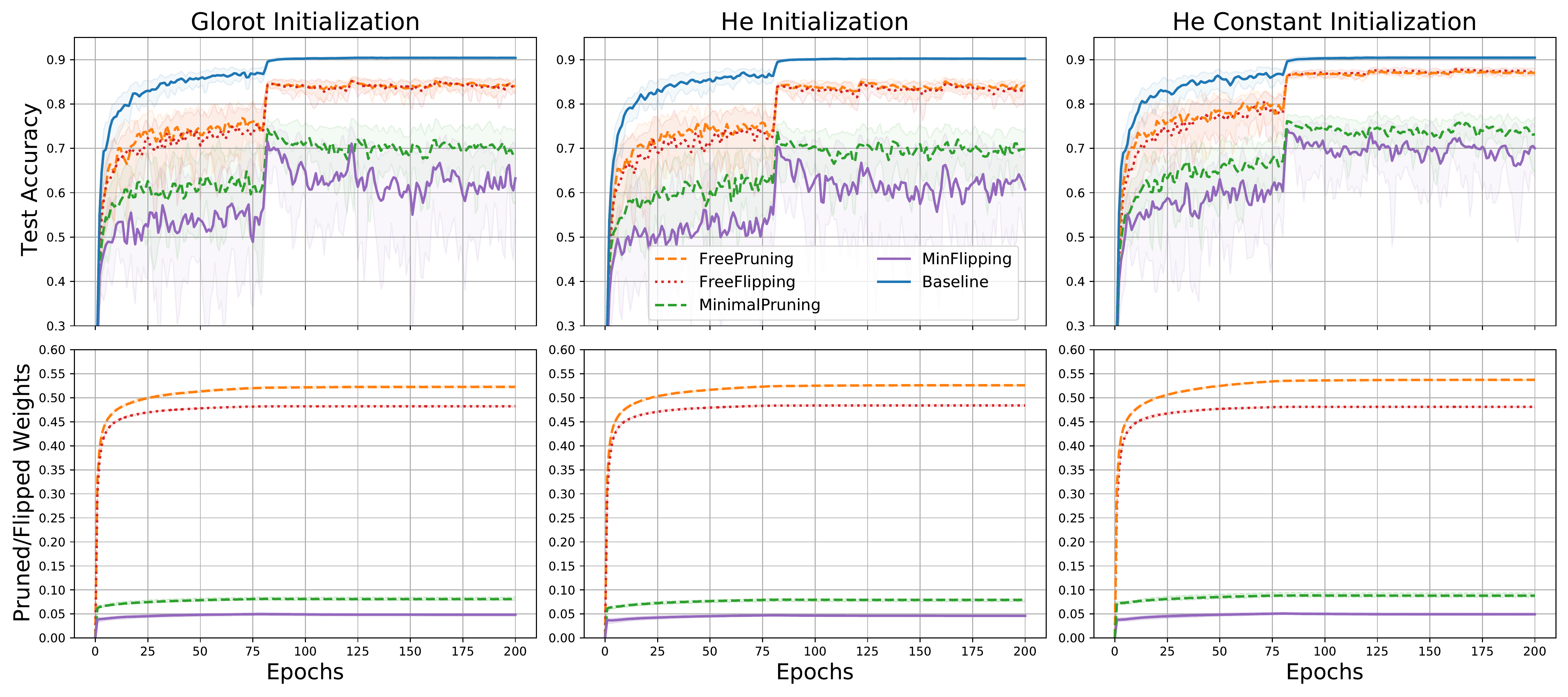}
		\caption{Pruning and sign flipping with ResNet-18 on CIFAR10 with different weight initializations.}
		\label{fig:resnets}
	\end{figure}
	
	We have performed all previous experiments on CIFAR-10 also on a convolutional network with residual connections. The training procedure is different than for the other networks: we augment the training images by random horizontal flipping, we use batch normalization layers and we vary the learning rate. We train the network for 200 epochs, where for the first 80 epochs the learning rate is $10^{-3}$, after which it is divided by 10 every 40 epochs. We have found that the regularization term in equation \ref{custom_loss} has to be multiplied by 0.2 in order to achieve a similar pruning/flipping rate.
	Figure \ref{fig:resnets} shows the performance of the free/minimal pruning and sign flipping algorithms compared to the baseline technique. In this case free pruning and free flipping perform almost exactly the same, being about 3 percentage points below the baseline accuracy. Minimal pruning and flipping are lower than the baseline but the number of pruned/flipped weights is very small for both. 
	
	\begin{table}
		\caption{Classification accuracies obtained by all training techniques. We indicate the best accuracy averaged over 5 runs for each experimental setup. This corresponds to the maximum of the	accuracy curves in all figures, excluding the uncertainty bands.}
		\centering
		\begin{tabular}[h]{cccccc}
			\toprule
			&  \textbf{LeNet} 	&  \textbf{Conv2} 	& \textbf{Conv4} & \textbf{Conv6} & \textbf{ResNet-18} \\ 
			\midrule
			Baseline 		&  97.96\% 	& \textbf{ 69.41\%}	& \textbf{76.04\%}	& 79.68\% & \textbf{90.45\%}\\ 
			\midrule
			Free pruning 	&  98.05\% 	&  68.71\%	& 74.68\%	& 78.22\% & 87.47\% \\ 
			\midrule
			Minimal pruning 	&  96.83\%	&  66.08\%	& 70.06\%	& 72.38\% & 76.13\%\\ 
			\midrule
			Free flipping	&  \textbf{98.14\%}	&  68.94\%	& 75.54\%	& \textbf{79.93\%} & 87.85\% \\
			\midrule 
			Minimal flipping	&  97.22\%	&  65.46\%	& 69.02\%	& 72.34\% & 74.34\% \\
			\bottomrule
		\end{tabular}
		\label{tab:results_summary}
	\end{table}

	\section{Conclusion and discussion}
	
	We have presented a simple and straightforward method for training the connectivity
	graph of a randomly initialized neural network directly through back-propagation, 
	without ever training the weights, even if the weights have constant magnitude. With our algorithms we can either switch on and
	off the connections between neurons or flip their signs. Our novel methods do not need hyperparameters defining cutoff thresholds, which removes the need of searching optimal values for such hyperparameters.
	Both methods yield very good results compared to training all weights of a network, in some cases even outperforming it, with free flipping of connection signs.
	We have also shown that it is possible to train a neural network connectivity graph
	essentially
	with a single weight which can be absorbed in the input data at training time, while
	at inference time the weight becomes irrelevant to the classification accuracy.
	We achieved good results even
	when the weights are drawn from very skewed distributions.
	
	It is not clear why deep neural networks generalize so well despite having far more trainable model parameters than the number of
	samples they are trained on \cite{ZhangBHRV17}. According to the Minimum Description Length principle \cite{grunwald2007minimum}, generalization capacity is correlated with a short size of the computer program performing the classification. In the case of a neural network, the program includes the actual computational steps needed for running the inference as well as any needed parameters or subprograms for generating these parameters. Our results contribute to finding ways of reducing the program length of deep neural networks. As also mentioned in \cite{LTH_uber}, if a network uses randomly initialized, untrained weights, then the weights do not need to be stored and may be represented as just a random number generator and its seed. In such a case, just a representation of the connectivity mask has to be integrated in the program. With our sign flipping training, using constant weights, the weights do not have to be represented at all, and just the filter $\mathbf{F}$ of one bit signs has to be represented in the program. With our minimal pruning or minimal flipping training procedures, we can generate sparse masks or filters, which also have compressed representations. Jointly with previous results that show that, through pruning, the number of weights in a classical neural network can be reduced significantly without loss of performance \cite{LTH_original}, our results suggest that classical deep neural networks are indeed over-parameterized, and that, through pruning or through alternative training methods like the sign flipping introduced here, their program size can be reduced to an effective, shorter one, which may explain their generalization power.    
	

	\section{Acknowledgements}
	This work was supported by the European Regional Development Fund and the Romanian Government through the Competitiveness Operational Programme 2014--2020, project ID P\_37\_679, MySMIS code 103319, contract no. 157/16.12.2016.

	\bibliographystyle{abbrv}

\begin{thebibliography}{10}
		
		\bibitem{STE}
		Y.~Bengio.
		\newblock Estimating or propagating gradients through stochastic neurons.
		\newblock {\em arXiv}, 1305.2982, 2013.
		
		\bibitem{Dai_2019}
		X.~Dai, H.~Yin, and N.~K. Jha.
		\newblock {NeST}: A neural network synthesis tool based on a grow-and-prune
		paradigm.
		\newblock {\em IEEE Transactions on Computers}, 68(10):1487–1497, 2019.
		
		\bibitem{LTH_original}
		J.~Frankle and M.~Carbin.
		\newblock The lottery ticket hypothesis: Finding sparse, trainable neural
		networks.
		\newblock In {\em International Conference on Learning Representations (ICLR)},
		2019.
		
		\bibitem{Glorot2010UnderstandingTD}
		X.~Glorot and Y.~Bengio.
		\newblock Understanding the difficulty of training deep feedforward neural
		networks.
		\newblock In {\em AISTATS}, 2010.
		
		\bibitem{grunwald2007minimum}
		P.~D. Gr{\"u}nwald.
		\newblock {\em The minimum description length principle}.
		\newblock MIT Press, 2007.
		
		\bibitem{resnet}
		K.~He, X.~Zhang, S.~Ren, and J.~Sun.
		\newblock Deep residual learning for image recognition.
		\newblock {\em CoRR}, abs/1512.03385, 2015.
		
		\bibitem{He2015DelvingDI}
		K.~He, X.~Zhang, S.~Ren, and J.~Sun.
		\newblock Delving deep into rectifiers: Surpassing human-level performance on
		{ImageNet} classification.
		\newblock {\em 2015 IEEE International Conference on Computer Vision (ICCV)},
		pages 1026--1034, 2015.
		
		\bibitem{hinton_STE}
		G.~Hinton.
		\newblock Neural networks for machine learning. {Coursera} video lectures,
		2012.
		
		\bibitem{adam}
		D.~P. Kingma and J.~Ba.
		\newblock Adam: {A} method for stochastic optimization.
		\newblock {\em CoRR}, abs/1412.6980, 2014.
		
		\bibitem{lenet300}
		Y.~LeCun, L.~Bottou, Y.~Bengio, and P.~Haffner.
		\newblock Gradient-based learning applied to document recognition.
		\newblock {\em Proceedings of the IEEE}, 86(11):2278--2324, 1998.
		
		\bibitem{mnist}
		Y.~LeCun and C.~Cortes.
		\newblock {MNIST} handwritten digit database.
		\newblock 2010.
		
		\bibitem{lee2018snip}
		N.~Lee, T.~Ajanthan, and P.~H.~S. Torr.
		\newblock {SNIP}: Single-shot network pruning based on connection sensitivity.
		\newblock In {\em International Conference on Learning Representations
			({ICLR})}, 2019.
		
		\bibitem{liu2020dynamic}
		J.~Liu, Z.~Xu, R.~Shi, R.~C.~C. Cheung, and H.~K.~H. So.
		\newblock Dynamic sparse training: Find efficient sparse network from scratch
		with trainable masked layers.
		\newblock In {\em International Conference on Learning Representations
			({ICLR})}, 2020.
		
		\bibitem{mostafa2019parameter}
		H.~Mostafa and X.~Wang.
		\newblock Parameter efficient training of deep convolutional neural networks by
		dynamic sparse reparameterization.
		\newblock In {\em International Conference on Learning Representations
			({ICLR})}, 2019.
		
		\bibitem{ramanujan2019whats}
		V.~Ramanujan, M.~Wortsman, A.~Kembhavi, A.~Farhadi, and M.~Rastegari.
		\newblock What's hidden in a randomly weighted neural network?
		\newblock {\em arXiv}, 1911.13299, 2019.
		
		\bibitem{vgg}
		K.~Simonyan and A.~Zisserman.
		\newblock Very deep convolutional networks for large-scale image recognition.
		\newblock In {\em International Conference on Learning Representations (ICLR)},
		2015.
		
		\bibitem{wang2020picking}
		C.~Wang, G.~Zhang, and R.~Grosse.
		\newblock Picking winning tickets before training by preserving gradient flow.
		\newblock In {\em International Conference on Learning Representations (ICLR)},
		2020.
		
		\bibitem{ZhangBHRV17}
		C.~Zhang, S.~Bengio, M.~Hardt, B.~Recht, and O.~Vinyals.
		\newblock Understanding deep learning requires rethinking generalization.
		\newblock In {\em International Conference on Learning Representations,
			({ICLR})}, 2017.
		
		\bibitem{LTH_uber}
		H.~Zhou, J.~Lan, R.~Liu, and J.~Yosinski.
		\newblock Deconstructing lottery tickets: Zeros, signs, and the supermask.
		\newblock In {\em Conference on Neural Information Processing Systems (NIPS)},
		2019.
		
	\end{thebibliography}

	\newpage
	\appendix
	\appendixpage
	
	\section{Experimental setup}\label{app_experimentalsetup}
	
	Table \ref{tab:table_archdetails} lists the architectures for all networks as well 
	as the optimizer and learning rates we used for each experiment and training/pruning
	methods.
	\setcounter{table}{0}
	\renewcommand{\thetable}{\thesection.\arabic{table}}

	\begin{table}[h]
		\footnotesize
		\caption{Neural networks architecture}
		\centering
		\begin{tabular}{c c c c c c }
			\toprule
			\makecell{\textbf{Model} \\\textbf{Dataset}} & \makecell{\textbf{LeNet}\\\textbf{MNIST}} &  \makecell{\textbf{Conv2} \\\textbf{CIFAR10}}& \makecell{\textbf{Conv4} \\\textbf{CIFAR10}} & \makecell{\textbf{Conv6} \\\textbf{CIFAR10}} & \makecell{\textbf{ResNet-18} \\\textbf{CIFAR10}}   \\
			\midrule
			Conv Layers & None & 2x64, pool & \makecell{2x64, pool \\ 2x128, pool}  &\makecell{2x64, pool \\ 2x128, pool\\ 2x256, pool} &\makecell{16, 3x[16, 16] \\ 3x[32, 32]\\ 3x[64, 64]} \\ 
			\midrule 
			FC Layers & 300, 100, 10 & 256, 256, 10 & 256, 256, 10  & 256, 256, 10 & avg-pool, 10 \\
			\midrule
			Batch size &  \multicolumn{4}{c}{25} & 64   \\
			\midrule
			\makecell{\textbf{Optimizer and} \\\textbf{Learning Rates}} &  \multicolumn{5}{c}{Adam}   \\
			\midrule
			\makecell{Baseline training} &  $  10^{-3}$  &  $2 \cdot 10^{-4}$ &  $3 \cdot 10^{-4}$ &   $3 \cdot 10^{-4}$ & $10^{-3} \rightarrow 10^{-6}$ \\		
			\midrule
			\makecell{Free pruning} &  $10^{-3}$ &  $3 \cdot 10^{-3}$  &  $3 \cdot 10^{-3}$  &  $3 \cdot 10^{-3}$  & $10^{-3} \rightarrow 10^{-6}$ \\
			\midrule
			\makecell{Minimal pruning} &  $10^{-3}$  &  $3 \cdot 10^{-3}$  &  $3 \cdot 10^{-3}$  &  $3 \cdot 10^{-3}$  & $10^{-3} \rightarrow 10^{-6}$ \\
			\midrule
			\makecell{Free flipping} &  $10^{-3}$  &  $5 \cdot 10^{-4}$ &  $5 \cdot 10^{-4}$ &  $5 \cdot 10^{-4}$ & $10^{-3} \rightarrow 10^{-6}$ \\
			\midrule
			\makecell{Minimal flipping} &  $10^{-3}$  &  $5 \cdot 10^{-4}$ &  $5 \cdot 10^{-4}$ &  $5 \cdot 10^{-4}$  & $10^{-3} \rightarrow 10^{-6}$\\
			\bottomrule 
		\end{tabular}
		\label{tab:table_archdetails}
	\end{table}

	\section{Forward propagation}\label{sumofpaths_derivation}
	\setcounter{figure}{0}
	\renewcommand\thefigure{\thesection.\arabic{figure}}

	\begin{figure}[h]
		\centering
		\includegraphics[width=.9\linewidth]{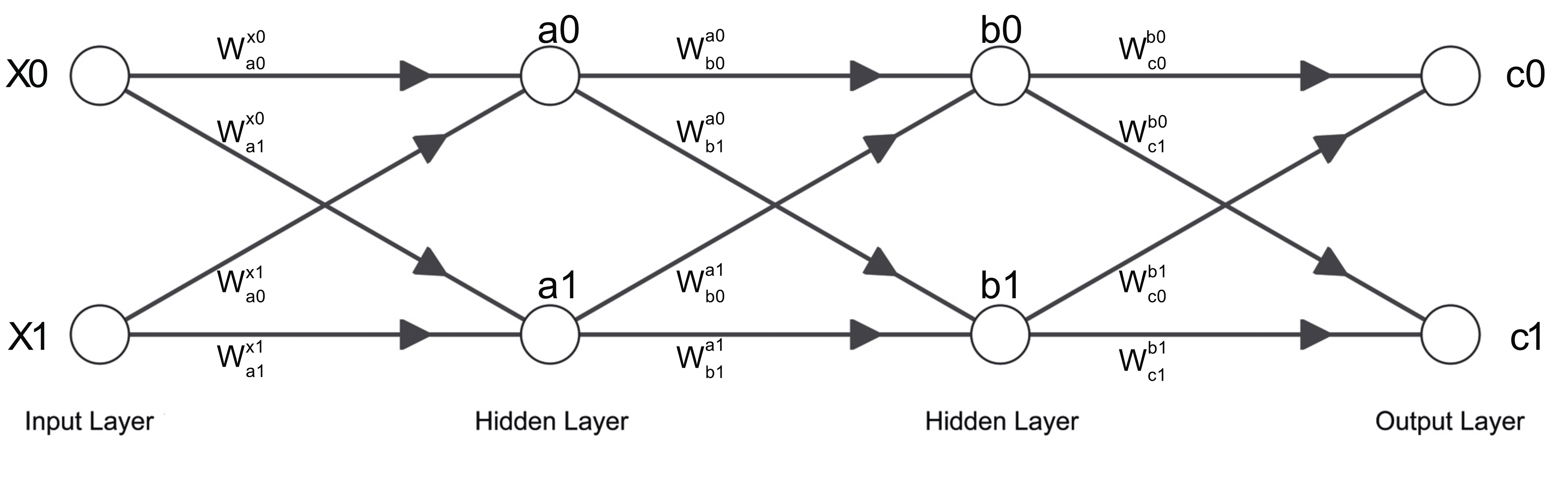}
		\caption{Simple neural network without bias nodes.}
		\label{fig:nn_1_1}
	\end{figure}
	
	Figure \ref{fig:nn_1_1} illustrates a simple neural network with two hidden layers, each with two neurons and no bias. We are interested in the output of every node in a forward propagation step.
	The output of each node is written as:
	
	\begin{align}\label{eq:nodes}
	a_0&=\sigma\left(x_{0}W_{a_0}^{x_0} +x_{1}W_{a_0}^{x_1}\right) &  b_0&=\sigma\left(a_{0}W_{b_0}^{a_0} +a_{1}W_{b_0}^{a_1}\right)  & c_0&=b_{0}W_{c_0}^{b_0} +b_{1}W_{c_0}^{b_1} \\
	a_1&=\sigma\left(x_{0}W_{a_1}^{x_0} +x_{1}W_{a_1}^{x_1}\right) &  b_1&=\sigma\left(a_{0}W_{b_1}^{a_0} +a_{1}W_{b_1}^{a_1}\right)  & c_1&=b_{0}W_{c_1}^{b_0} +b_{1}W_{c_1}^{b_1}
	\end{align}
	
	where $\sigma(x)$ is a non-linear activation function.
	Note that for the output nodes $c_i$ we do not apply an activation function, we are
	only interested in their values. Choosing $\sigma(x)=\max(0,x)$ we have the convenient 
	property that $\sigma(x)=xH(x)$, where  
	$H(x) = \begin{cases}
	0, & x \leq 0 \\
	1, & x>0
	\end{cases}\, $ is the step function. \\
	Thus we can rewrite the output for the nodes $a_i$ and $b_i$ as:
	
	\begin{align}\label{eq:a_0_a1}
	a_0&=\left(x_{0}W_{a_0}^{x_0} +x_{1}W_{a_0}^{x_1}\right)H_{a_0}  &  a_1&=\left(x_{0}W_{a_1}^{x_0} +x_{1}W_{a_1}^{x_1}\right)H_{a_1} \\
	b_0&=\left(a_{0}W_{b_0}^{a_0} +a_{1}W_{b_0}^{a_1}\right) H_{b_0} &  b_1&=\left(a_{0}W_{b_1}^{a_0} +a_{1}W_{b_1}^{a_1}\right)H_{b_1}
	\end{align}
	
	where $H_{a_i}$ and $H_{b_i}$ are short notations for the step function applied at $a_i$ and $b_i$ in order to
	avoid rewriting each time the long expression for the argument of $H$. 
	Replacing $a_i$ in $b_i$ we obtain:

	\begin{align}\label{eq:b_0_b1_step1}
	b_0&=\left[\left(x_{0}W_{a_0}^{x_0} +x_{1}W_{a_0}^{x_1}\right)H_{a_0}W_{b_0}^{a_0} +\left(x_{0}W_{a_1}^{x_0} +x_{1}W_{a_1}^{x_1}\right)H_{a_1}W_{b_0}^{a_1}\right] H_{b_0} &  \\ 
	b_1&=\left[\left(x_{0}W_{a_0}^{x_0} +x_{1}W_{a_0}^{x_1}\right)H_{a_0}W_{b_1}^{a_0} +\left(x_{0}W_{a_1}^{x_0} +x_{1}W_{a_1}^{x_1}\right)H_{a_1}W_{b_1}^{a_1}\right]H_{b_1}
	\end{align}
	
	We can regroup terms and give $x_0$ and $x_1$ common factors:

	\begin{align}\label{eq:b_0_b1}
	b_0&=\left[ x_{0}\left(W_{a_0}^{x_0}W_{b_0}^{a_0}H_{a_0} +W_{a_1}^{x_0}W_{b_b}^{a_1}H_{a_1} \right) +x_{1} \left(W_{a_0}^{x_1}W_{b_0}^{a_0}H_{a_0} +W_{a_1}^{x_1}W_{b_0}^{a_1}H_{a1}  \right) \right]H_{b_0} \\
	b_1&=\left[x_{0} \left(W_{a_0}^{x_0}W_{b_1}^{a_0}H_{a_0} +W_{a_1}^{x_0}W_{b_1}^{a_1}H_{a_1} \right) +x_{1} \left(W_{a_0}^{x_1}W_{b_1}^{a_0}H_{a_0}+W_{a_1}^{x_1}W_{b_1}^{a_1}H_{a_1}  \right) \right]H_{b_1}
	\end{align}
	
	Replacing $b_i$ in $c_0$, multiplying and giving $x_0$ and $x_1$ common factors results in:
	
	\begin{alignat}{2}\label{eq_appendix:c0_final}
	c_{0} = & x_0 \Big[ & W_{a_0}^{x_0}W_{b_0}^{a_0}W_{c_0}^{b_0}H_{a_0}H_{b_0} &+W_{a_1}^{x_0}W_{b_0}^{a_1}W_{c_0}^{b_0}H_{a_1}H_{b_0}  +\\ \nonumber
	&& W_{a_0}^{x_0}W_{b_1}^{a_0}W_{c_0}^{b_1}H_{a_0}H_{b_1} &+W_{a_1}^{x_0}W_{b_1}^{a_1}W_{c_0}^{b_1}H_{a_1}H_{b_1} \Big] +\\ \nonumber
	& 	  x_1 \Big[ & W_{a_0}^{x_1}W_{b_0}^{a_0}W_{c_0}^{b_0}H_{a_0}H_{b_0} &+W_{a_1}^{x_1}W_{b_0}^{a_1}W_{c_0}^{b_0}H_{a_1}H_{b_0}  +\\ \nonumber
	&& W_{a_0}^{x_1}W_{b_1}^{a_0}W_{c_0}^{b_1}H_{a_0}H_{b_1} &+W_{a_1}^{x_1}W_{b_1}^{a_1}W_{c_0}^{b_1}H_{a_1}H_{b_1} \Big]  
	\end{alignat}

	\newpage
	\section{Full comparison between algorithms}\label{fulcomparison}
	\setcounter{figure}{0}
	\renewcommand\thefigure{\thesection.\arabic{figure}}
	The figures in this section indicate the performance of the free/minimal pruning and flipping algorithms compared to the baseline for all networks. 
	
	\begin{figure}[h]
		\centering
		\includegraphics[width=1\linewidth]{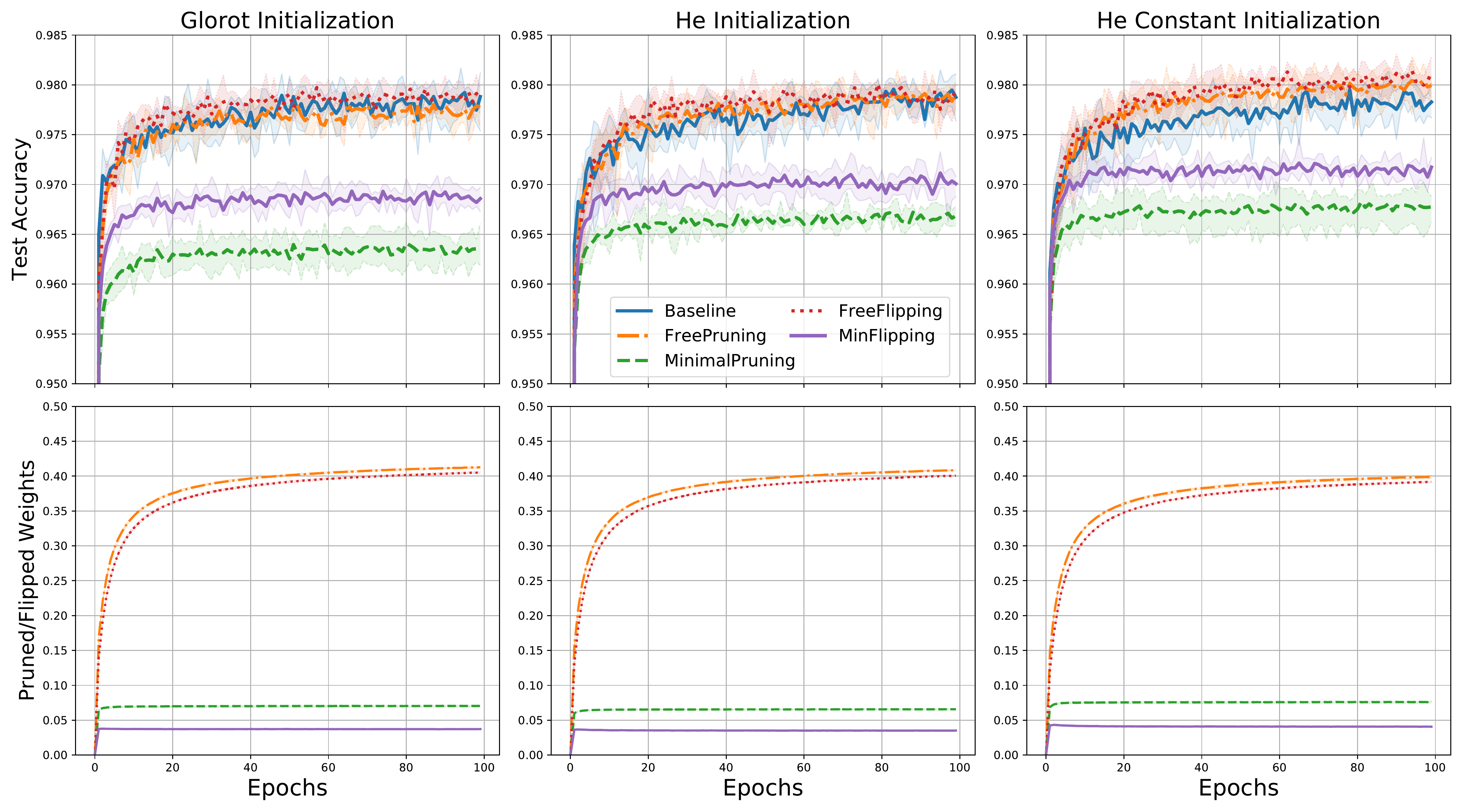}
		\caption{LeNet on MNIST with different initializations.}
		\label{fig:0baselinesfreepruningminpruningsignflipminflip}
	\end{figure}
	
	\begin{figure}[h]
		\centering
		\includegraphics[width=1\linewidth]{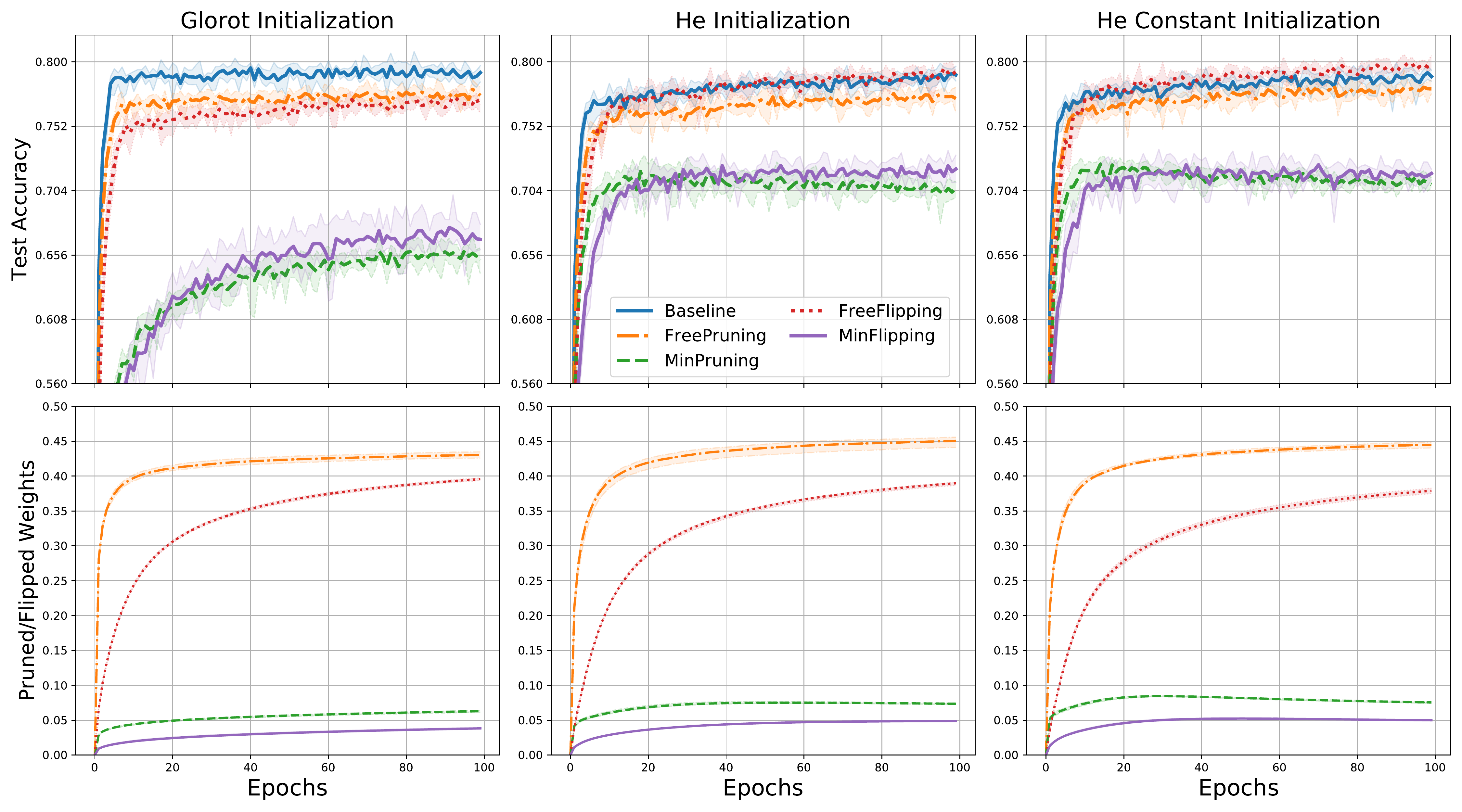}
		\caption{Conv6 on CIFAR10 with different weight initializations.}
		\label{fig:0_Acc_Baseline_FreePruning_MinPruning_SignFlip_MinFlip_C6}
	\end{figure}

	\begin{figure}[h]
		\centering
		\includegraphics[width=1\linewidth]{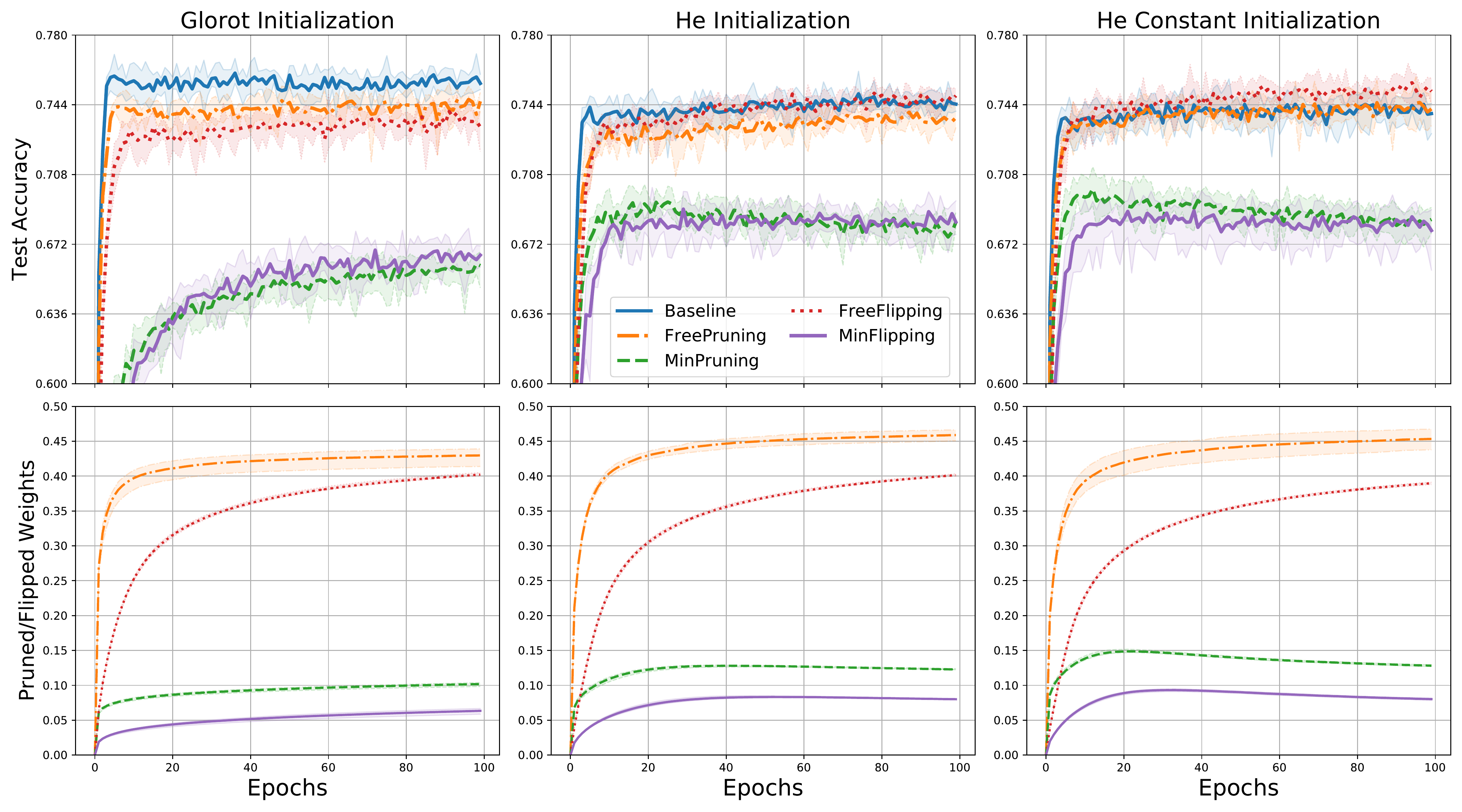}
		\caption{Conv4 on CIFAR10 with different weight initializations.}
		\label{fig:0_Acc_Baseline_FreePruning_MinPruning_SignFlip_MinFlip_C4}
	\end{figure}
	
	\begin{figure}[h]
		\centering
		\includegraphics[width=1\linewidth]{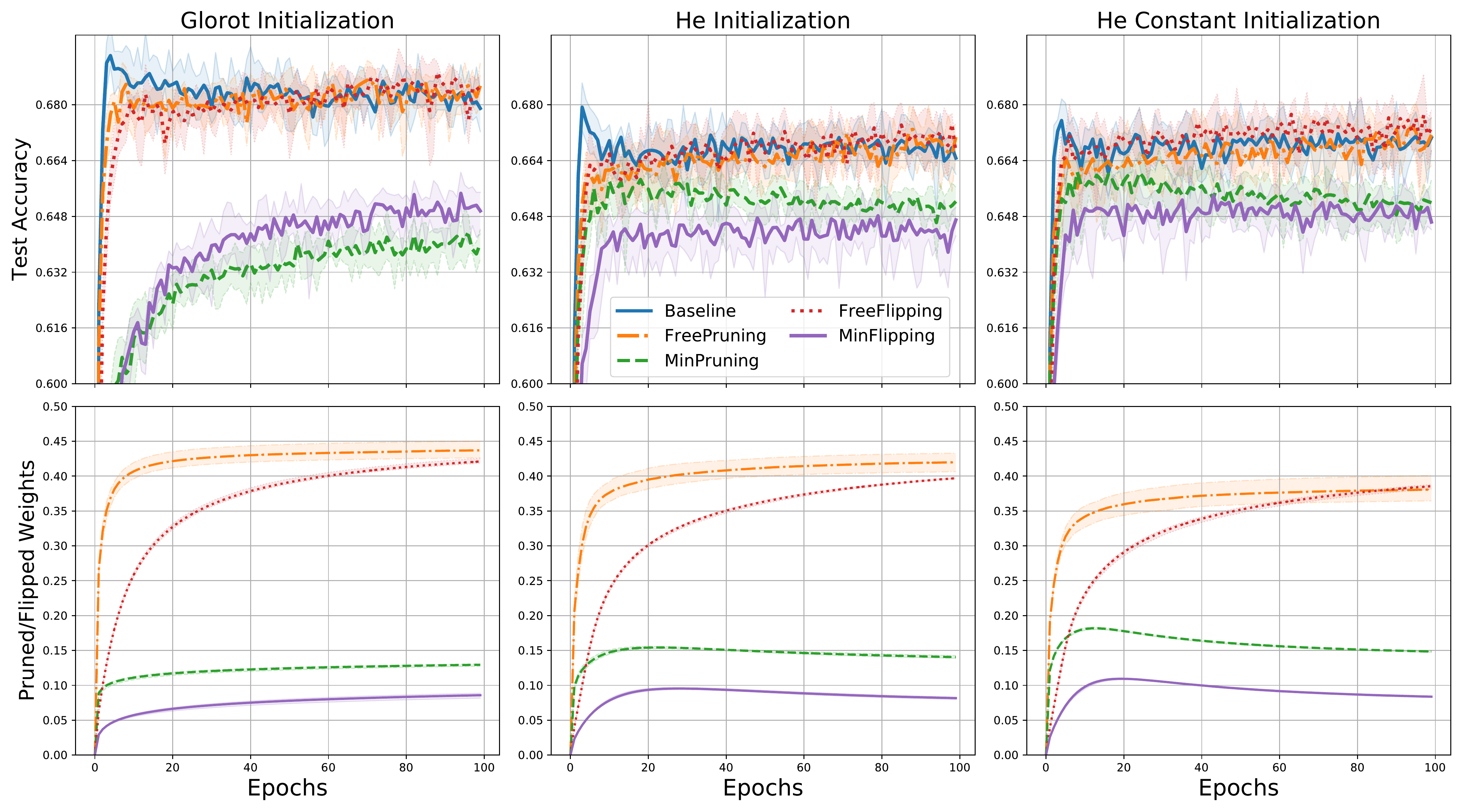}
		\caption{Conv2 on CIFAR10 with different weight initializations.}
		\label{fig:0_Acc_Baseline_FreePruning_MinPruning_SignFlip_MinFlip_C2}
	\end{figure}

\end{document}